\documentclass[letterpaper]{article} 
\usepackage{aaai2026}  
\usepackage{times}  
\usepackage{helvet}  
\usepackage{courier}  
\usepackage[hyphens]{url}  
\usepackage{graphicx} 
\usepackage{natbib}  
\usepackage{caption} 
\usepackage{graphicx}
\usepackage{subfig}
\usepackage{amsfonts}
\usepackage{multicol}
\usepackage{soul}
\usepackage{tabularray}

\usepackage{algorithm}
\usepackage{algorithmic}

\usepackage{newfloat}
\usepackage{listings}
\usepackage{amsmath}
\usepackage{cleveref}
\usepackage{bbding}

\usepackage{newfloat}
\usepackage{listings}
\DeclareCaptionStyle{ruled}{labelfont=normalfont,labelsep=colon,strut=off} 
\floatstyle{ruled}
\newfloat{listing}{tb}{lst}{}
\floatname{listing}{Listing}
\title{Towards Efficient Low-rate Image Compression with Frequency-aware 
\\ Diffusion Prior Refinement}
\author{
    Yichong Xia\equalcontrib \textsuperscript{\rm 1,2}, 
     Yimin Zhou\equalcontrib \textsuperscript{\rm 1}, Jinpeng Wang \textsuperscript{\rm 3},
  Bin Chen \thanks{Corresponding author(s)}\textsuperscript{\rm 3}
}
\affiliations{
    \textsuperscript{\rm 1}Tsinghua Shenzhen International Graduate School, Tsinghua University\\

     \textsuperscript{\rm 2}Peng Cheng Laboratory\\
     \textsuperscript{\rm 3}Harbin Institute of Technology, Shenzhen \\
    {xiayc23,zhouym24,wjp20}@mails.tsinghua.edu.cn, chenbin2021@hit.edu.cn
}

\usepackage{bibentry}

\begin{document}

\maketitle

\begin{abstract}
Recent advancements in diffusion-based generative priors have enabled visually plausible image compression at extremely low bit rates. However, existing approaches suffer from slow sampling processes and suboptimal bit allocation due to fragmented training paradigms. In this work, we propose Accelerate \textbf{Diff}usion-based Image Compression via \textbf{C}onsistency Prior \textbf{R}efinement (DiffCR), a novel compression framework for efficient and high-fidelity image reconstruction. At the heart of DiffCR is a Frequency-aware Skip Estimation (FaSE) module that refines the $\epsilon$-prediction prior from a pre-trained latent diffusion model and aligns it with compressed latents at different timesteps via Frequency Decoupling Attention (FDA). Furthermore, a lightweight consistency estimator enables fast \textbf{two-step decoding} by preserving the semantic trajectory of diffusion sampling. Without updating the backbone diffusion model, DiffCR achieves substantial bitrate savings (27.2\% BD-rate (LPIPS) and 65.1\%  BD-rate (PSNR)) and over $10\times$ speed-up compared to SOTA diffusion-based compression baselines. 
\end{abstract}


\section{Introduction}

In recent years, remarkable progress has been made in image compression technology. However, compressing high-definition images at low bit-rate still poses significant challenges.
Traditional image compression standards \citep{bpgbellard, jpeg} rely on manually designed transform methods. Nevertheless, in low bit-rate ($\leq0.1$ bpp) scenarios, prominent block artifacts and color distortion issues occur. Although learning-based end - to - end neural compression models \citep{cheng2020,balle17,balle18,he2022elic} outperform traditional methods in terms of performance, their distortion-driven optimization objectives still lead to substantial loss of texture and details in the reconstructed images when bandwidth is limited. \citep{blau2019rethinking,agustsson2023multi} summarizes this phenomenon as a triple trade-off between bit rate, distortion, and realism, revealing the inherent contradiction that extreme compression inevitably entails semantic information loss and a decline in perceptual quality.

\begin{figure}[t!]
\centering
\includegraphics[width=1.0\linewidth]{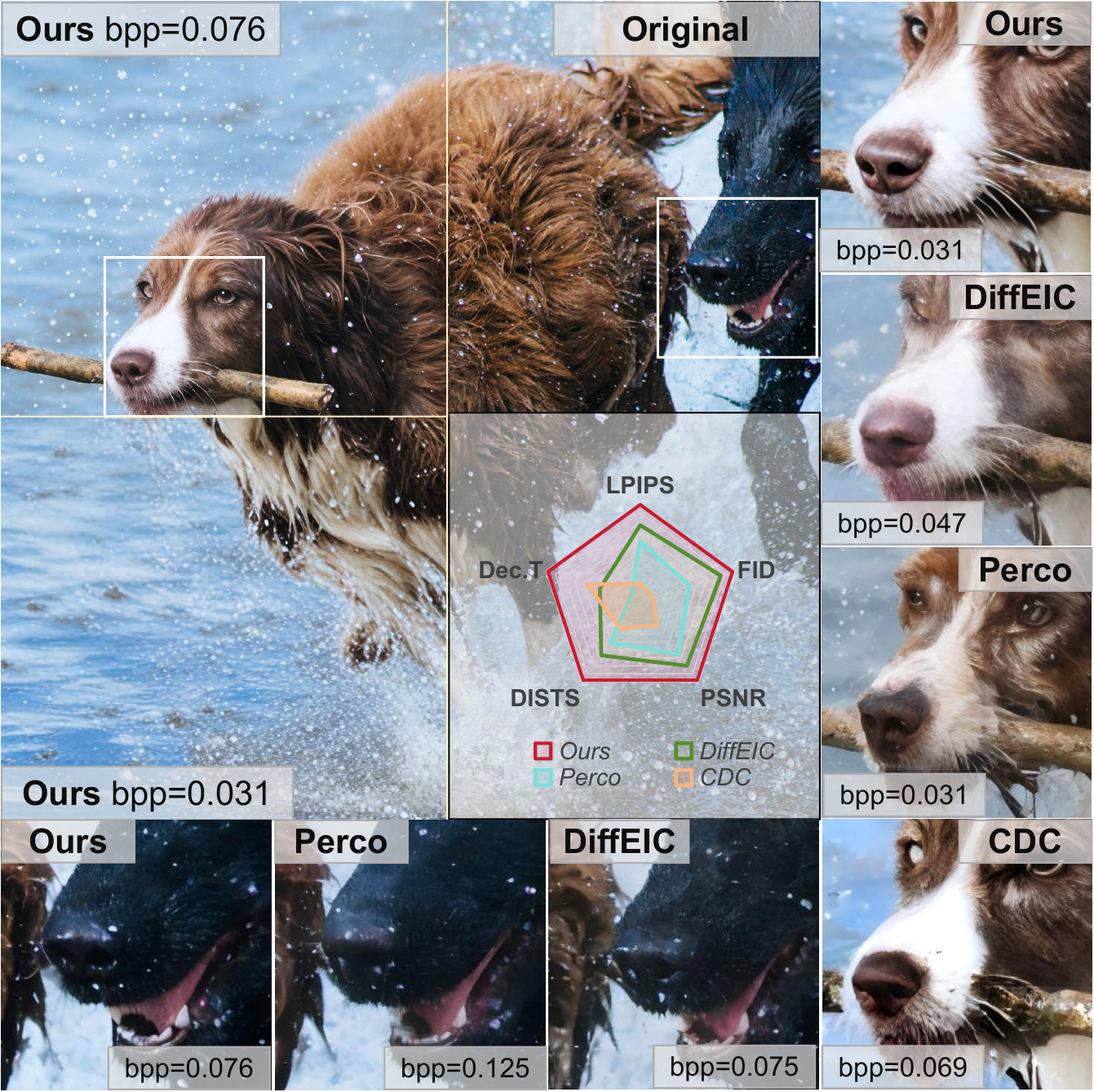}
\caption{Qualitative and quantitative comparison between DiffEIC, Perco, HiFiC, MS-ILLM, CDC, and our proposed approach. DiffCR (Ours) not only possesses realism but also faithfully restores the details of the original image. The radar chart in the bottom right corner illustrates the performance comparison among various methods on the CLIC20 dataset. \emph{Best viewed when zoomed in.}}
\label{fig:bigpic}
\end{figure}

To address this issue, recent studies \citep{HIFIC,ILLM,lu2024hybridflow,qin2023perceptual} have introduced generative adversarial networks (GAN) to optimize perceptual quality. Based on this, \citep{lee2024neural} introduces semantic information at both the encoding and decoding ends to enhance the image restoration effect. Constrained by the GAN structure and the single perceptual loss, these methods still have considerable room for improvement in statistical fidelity. Meanwhile, some 3D generative methods based on neural representation and Gaussian splatting \citep{liuexploration,liu2025efficient} have also been applied to lossy compression tasks for videos and images.

In contrast, diffusion models \citep{dhariwal2021diffusion,LDM} demonstrate superior generative quality through a progressive denoising process. The statistical characteristics of their reconstructed images are closer to the distribution of natural images. \citep{CDC,hoogeboom2023high} first applied the  DDPM \citep{DDPM} framework to the field of image compression and achieved breakthrough progress. However, this method requires the complete training of diffusion model components and has problems such as high computational costs and generalization being limited by the training data.

More recently, several studies based on Latent Diffusion Models (LDMs) \citep{LDM} have introduced novel concepts to overcome the issues present in the GAN framework.
For instance, \citep{lei2023text+} endeavors to manipulate the pre-trained LDM by encoding sketch and text semantics. Utilize the sampling process of the diffusion model for image decoding. Given the difficulties in jointly optimizing semantic embeddings and the transfer representation for compression objectives, they have dedicated significant resources to iterative semantic embedding and alignment.
\citep{perco} makes use of a fully-trained LDM and encodes conditional images with a trainable codebook. However, this approach has two drawbacks. Firstly, it demands careful optimization of the model on a dataset consisting of millions of images. Secondly, the independent training of the compressor and the diffusion model diminishes the learning capacity of the compressor.
\citep{li2024towards} utilizes a pre-trained foundation LDM. Nevertheless, the joint training method it employs makes it arduous for the compressor to learn compact representations by capitalizing on the diffusion prior.
\citep{ke2025ultra} harnesses Multimodal Large Language Models (MLLM) to acquire high-quality text semantics and attains remarkable performance improvements at ultra low bit-rate. Nevertheless, this comes at the cost of extremely high encoding latency.

In general, existing image compression schemes based on diffusion suffer from two key deficiencies. (1) Suboptimal bit allocation due to training paradigms and latent misalignment. In current research, the training of the compressor and diffusion model is either conducted separately \citep{CDC,perco} or the diffusion loss and compressor loss are misaligned \citep{li2024towards, ke2025ultra}. \citep{xiadiffpc} attempts have been made to employ proxy losses to enable the compressor to capture diffusion priors during training, yet the most compact compression representation remains elusive.
(2) Methods based on diffusion rely on standard diffusion models and encounter challenges in sampling as they require multiple iterations to generate realistic image outputs. This not only leads to significant decoding delays but also hinders their seamless integration with perceptual losses applied to the final image output, consequently diminishing decoding fidelity.

    To address the aforementioned critical issues, we introduce the Accelerate \textbf{Diff}usion-based Image Compression via \textbf{C}onsistency Prior \textbf{R}efinement (DiffCR). Reflecting upon the training paradigm of $\epsilon$-Prediction in the fundamental diffusion model, we devise a Consistency Refinement Estimator (CRE) to actively correct and enhance diffusion priors, aligning the model's focus on optimizing both the realism of generated images and the quest for compact compression representations. Specifically, we craft the Frequency-aware Skip Estimation (FaSE) to align diffusion priors with compressed latents at different time steps and employ Frequency Decoupling Attention (FDA) to capture contextual relationships between the two at various frequencies. Furthermore, we maintain the consistency of FaSE sampling trajectories with original ODE trajectories through consistency losses. Unlike conventional consistency distillation methods, our approach achieves high-fidelity decoding in just two steps with only $1/100$ of the parameters of the original denoising network (8M vs.800M). Furthermore, DiffCR leverages mixed semantic flows to further fortify and stabilize the generated results. Benefitting from these characteristics, DiffCR decodes highly realistic images at extremely low bitrates ($\leq$0.05 bpp) in a mere 0.4 seconds, as depicted in \cref{fig:bigpic}.

Our contributions can be summarized as follows:
\begin{itemize}
    \item We introduce the Frequency-aware Skip Estimation to refine (FaSE) and align pre-trained diffusion priors with compressed latents. This not only facilitates the compressor in acquiring more compact compression representations but also allows our model to decode high-quality results in just two steps through consistency constraints.

    \item We devise the Frequency Decoupling Attention (FDA) to synchronize diffusion priors generated at different time steps with compressed latents. FDA decouples the two in the Fourier domain and modulates them across different time steps using a temporal mask. 
    \item Building upon these key points, we propose DiffCR. DiffCR not only enhances encoding efficiency but also utilizes the semantic branch to stabilize and enhance decoding performance. We validate DiffCR on multiple high-quality datasets, demonstrating its superior perceptual performance. Compared to a state-of-the-art compression approach  \citep{li2024towards}, which is also based on SD 2.1, DiffCR achieves savings in BD-rate of 27.2\% (LPIPS), 32.8\% (FID), and 65.1\% (PSNR).

\end{itemize}
\begin{figure*}[t]
\centering
\includegraphics[width=1\linewidth]{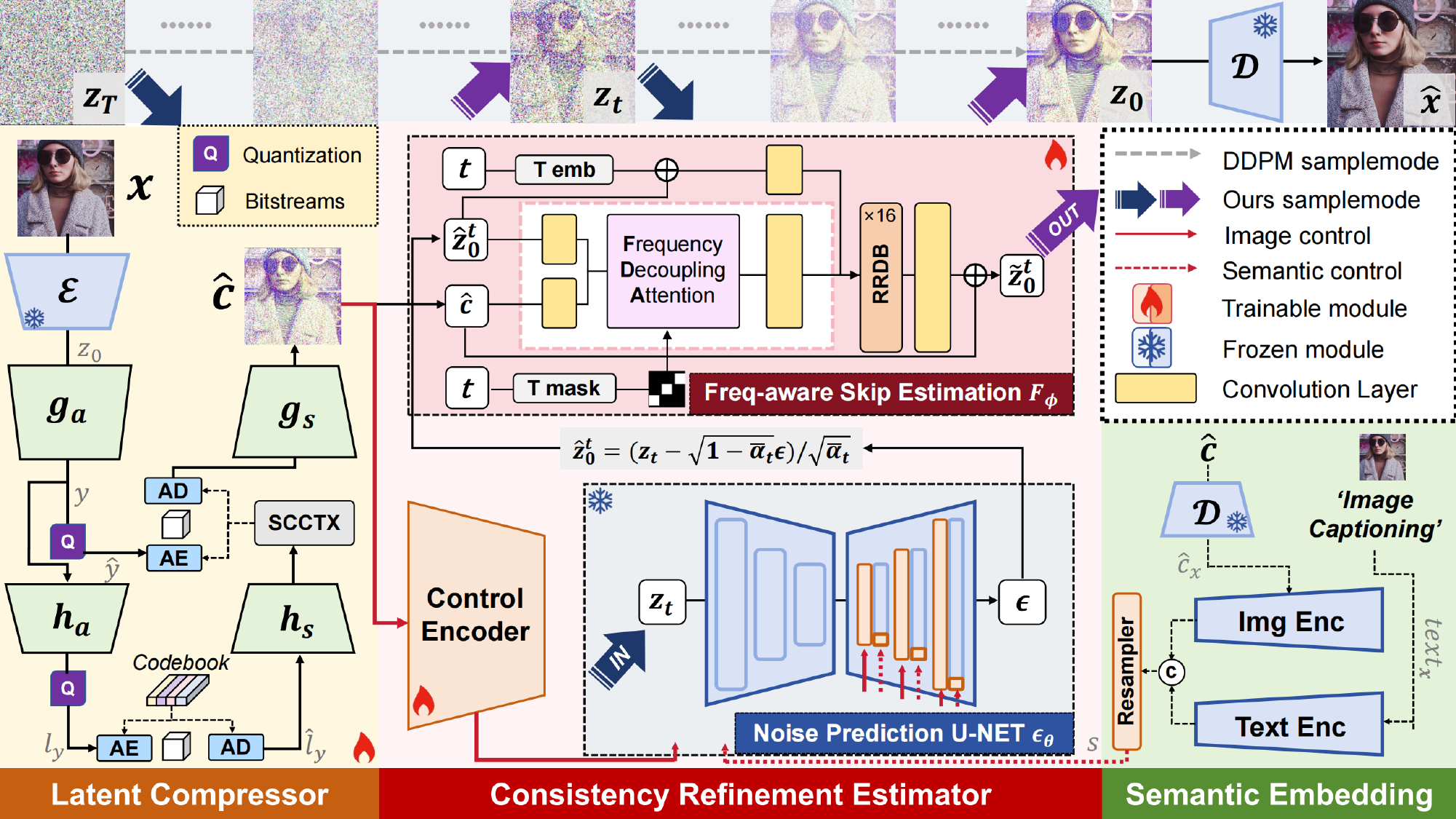}
\caption{Illustration of the proposed DiffCR. RRDB signifies Residual in Residual Dense Block \citep{wang2018esrgan}. DiffCR utilizes two branches to guide pre-trained denoising networks in generation and reinforces prior correction and refinement through the FaSE module, aligning it with the reconstruction objective of the compressor. Unlike the ordinary sampling process (depicted by the grey dashed line), DiffCR permits a two-step sampling approach (illustrated by the blue-purple arrows).}
\label{fig:pipeline}
\end{figure*}
\section{\textbf{Background}}
\subsection{Learned Latent Compression}
Learnable image compression \citep{balle17,balle18,minnen2018,he2022elic,cheng2020,qin2024mambavc} involves extracting image representations through an encoder, estimating the probability distribution of these representations using entropy models, and subsequently performing entropy coding. Following Shannon's information theory \citep{shannon1948mathematical}, the training paradigm can be unified into the following form:
\begin{align}
    \mathcal{L}=R_{\boldsymbol{x}}+\lambda D(\boldsymbol{x}, \mathcal{M}(\boldsymbol{x}))
\end{align}
Here, $\mathcal{M}(\cdot)$ represents the compressor, and $R_{\boldsymbol{x}}$ denotes the bitrate of the image $\boldsymbol{x}$. The function $D(\cdot)$ typically selects pixel-level distortions such as Mean Squared Error (MSE) or MS-SSIM \citep{ssim}. In estimating some generative compression tasks \citep{perco,mao2024extreme,li2024towards, CDC}, the compressor replaces encoding the image with a latent extracted by a pretrained analysis encoder. In this scenario, the training objective is replaced by:
\begin{align}
\label{eq:latent_c}
    \mathcal{L}=R_{\mathcal{E}(\boldsymbol{x})}+\lambda D(\mathcal{E}(\boldsymbol{x}), \mathcal{M}(\mathcal{E}(\boldsymbol{x})))
\end{align}
Here, $\mathcal{E}(\cdot)$ represents the pretrained analysis encoder. This approach eliminates the need for designing intricate feature extraction modules for the compressor, while ensuring that the decoded features reside in the encoding space of the pretrained generative model, thereby enhancing the quality of the final reconstructed image. Despite lightening the load on the compressor, this method introduces a potential inconsistency between distortions in the feature domain and distortions in the image domain, which could lead the compressor astray from the optimal rate-distortion trade-off solution.
\subsection{Latent Diffusion Models}
Diffusion models (DMs), or score-based generative models \citep{DDPM,songscore}, are a type of generative model that gradually injects Gaussian noise into the data and then generates samples from the noise through a reverse denoising process. To enable diffusion models to train on limited computing resources while preserving generation quality, Latent Diffusion Models (LDMs) \citep{LDM} encode an image $\boldsymbol{x}$ into a latent representation $\boldsymbol{z_0}$ via an encoder $\mathcal{E}$ and reconstruct it using a decoder $\mathcal{D}$.  

In LDMs' forward process, Gaussian noise is gradually added to the clean latent feature $\boldsymbol{z_0}$, with noise intensity at each step controlled by schedule $\beta_t$. This process is formulated as:
\begin{align}
\label{eq:eps}
\boldsymbol{z_t}=\sqrt{\bar{\alpha}_t} \boldsymbol{z_0}+\sqrt{1-\bar{\alpha}_t} \epsilon, \quad t \in\{1,2, \ldots, T\},
\end{align}
where $\epsilon \sim \mathcal{N}(0, \mathbf{I})$ is standard Gaussian noise. Here, $\alpha_t=1-\beta_t$ and $\bar{\alpha}_t=\prod_{i=1}^t \alpha_i$. As $t$ increases, the corrupted $\boldsymbol{z_t}$ progressively approximates a Gaussian distribution.
The reverse process of modeling LDMs is equivalent to training a noise prediction network $\boldsymbol{\epsilon}_{\theta}$ with the diffusion loss:
\begin{align}
\label{eq:dloss}
\mathcal{L}_{\text{diff}}=\mathbb{E}_{\boldsymbol{z_0}, t, \epsilon \sim \mathcal{N}(\mathbf{0}, \mathbf{I})}\left\|\boldsymbol{\epsilon}-\boldsymbol{\epsilon}_\theta\left(\boldsymbol{z_t}, \boldsymbol{c}, t\right)\right\|_2^2.
\end{align}
During inference, LDMs predict noise using the pre-trained denoising network $\boldsymbol{\epsilon}_\theta(\boldsymbol{z_t}, \boldsymbol{c}, t)$ with text condition $\boldsymbol{c}$, generating latent $\boldsymbol{z_{t-1}}$  to sequentially obtain the final latent $\boldsymbol{z_0}$
(As indicated by the grey dashed line in \cref{fig:pipeline}). Stable Diffusion, as pre-trained LDMs, have been shown to have priors that can be widely applied in low-level image tasks \citep{lin2023diffbir,wang2024exploiting,li2024diffusion}.
\subsection{Consistency Models}
Consistency Models (CMs) \citep{song2023consistency} use consistency mapping to directly map any point in an ODE trajectory back to its origin, enabling fast few-step generation. Formally, this mapping is defined as $\boldsymbol{f}:\left(\boldsymbol{x}_t, t\right) \longmapsto \boldsymbol{x}_\epsilon$ where $\epsilon$ is a fixed small positive number. A key property is self-consistency:
\begin{align}
\label{eq:cm1}
\boldsymbol{f}\left(\boldsymbol{x}_t, t\right)=\boldsymbol{f}\left(\boldsymbol{x}_{t^{\prime}}, t^{\prime}\right), \forall t, t^{\prime} \in[\epsilon, T] .
\end{align}
To ensure $\boldsymbol{f}_{\boldsymbol{\theta}}(\boldsymbol{x}, \epsilon)=\boldsymbol{x}$, the model is parameterized as:
\begin{align}
\label{eq:cm2}
\boldsymbol{f}_{\boldsymbol{\theta}}(\boldsymbol{x}, t)=c_{\text {skip }}(t) \boldsymbol{x}+c_{\text {out }}(t) \boldsymbol{F}_{\boldsymbol{\theta}}(\boldsymbol{x}, t),
\end{align}
where $c_{\text {skip }}(0)=1, c_{\text {out }}(0)=0$. During training, self-consistency is enforced using a target model $\boldsymbol{\theta}^{-}$—an exponential moving average (EMA) of $\boldsymbol{\theta}$: $\boldsymbol{\theta}^{-} \leftarrow \mu \boldsymbol{\theta}^{-}+(1-\mu) \boldsymbol{\theta}$. The consistency loss is:
\begin{align}
\begin{aligned}
\label{eq:cm3}
&\mathcal{L}\left(\boldsymbol{\theta}, \boldsymbol{\theta}^{-} ; \Phi\right)
=
\\
&\mathbb{E}_{\boldsymbol{x}, t}\left[d\left(\boldsymbol{f}_{\boldsymbol{\theta}}\left(\boldsymbol{x}_{t_{n+1}}, t_{n+1}\right), \boldsymbol{f}_{\boldsymbol{\theta}-}\left(\hat{\boldsymbol{x}}_{t_n}^{\Psi}, t_n\right)\right)\right].
\end{aligned}
\end{align}
Here, $d(\cdot, \cdot)$ is a distance metric. $\Psi$ denotes a one-step ODE solver estimating $\boldsymbol{x}_{t_n}$ from $\boldsymbol{x}_{t_{n+1}}$. \citep{lcm} applies this to pre-trained LDMs with skip-step scheduling, ensuring consistency between non-adjacent steps $t_{n+k} \rightarrow t_n$ instead of $t_{n+1} \rightarrow t_n$.

\section{Method}
\subsection{Overview Framework}
The \cref{fig:pipeline} illustrates the overall framework of DiffCR. For a detailed algorithmic flow, please refer to the appendix. The image $\boldsymbol{x}$ is first passed through the encoder $\mathcal{E}(\cdot)$ to obtain the latent representation $\boldsymbol{z_0}$ of the LDMs, which is then fed into the latent compressor for lossy compression, resulting in a bitstream and its decoded output $\boldsymbol{\hat{c}}$. $\boldsymbol{\hat{c}}$ guides the denoising network through the image control branch. Simultaneously, we mix the semantic information provided by the distorted image $\mathcal{D}(\boldsymbol{\hat{c}})$ and the textual semantics from $\boldsymbol{x}$, embedding them into the denoising network.

Next, DiffCR corrects and refines the output of the denoising network, $\boldsymbol{\epsilon}$, and image-level control $\boldsymbol{\hat{c}}$, through Frequency-aware Skip Estimation, mapping any input at any time ($\boldsymbol{\hat{z}^t_0}$) back to the origin status $\boldsymbol{z_0}$ ($\boldsymbol{\tilde{z}^t_0}$). Unlike the sampling path of DDPM, DiffCR allows for two-step fast sampling. The final sampling result is obtained through the decoder $\mathcal{D}(\cdot)$ to yield the ultimate decoded output $\boldsymbol{\hat{x}}$.
\begin{figure}[t]
\centering
\includegraphics[width=1\linewidth]{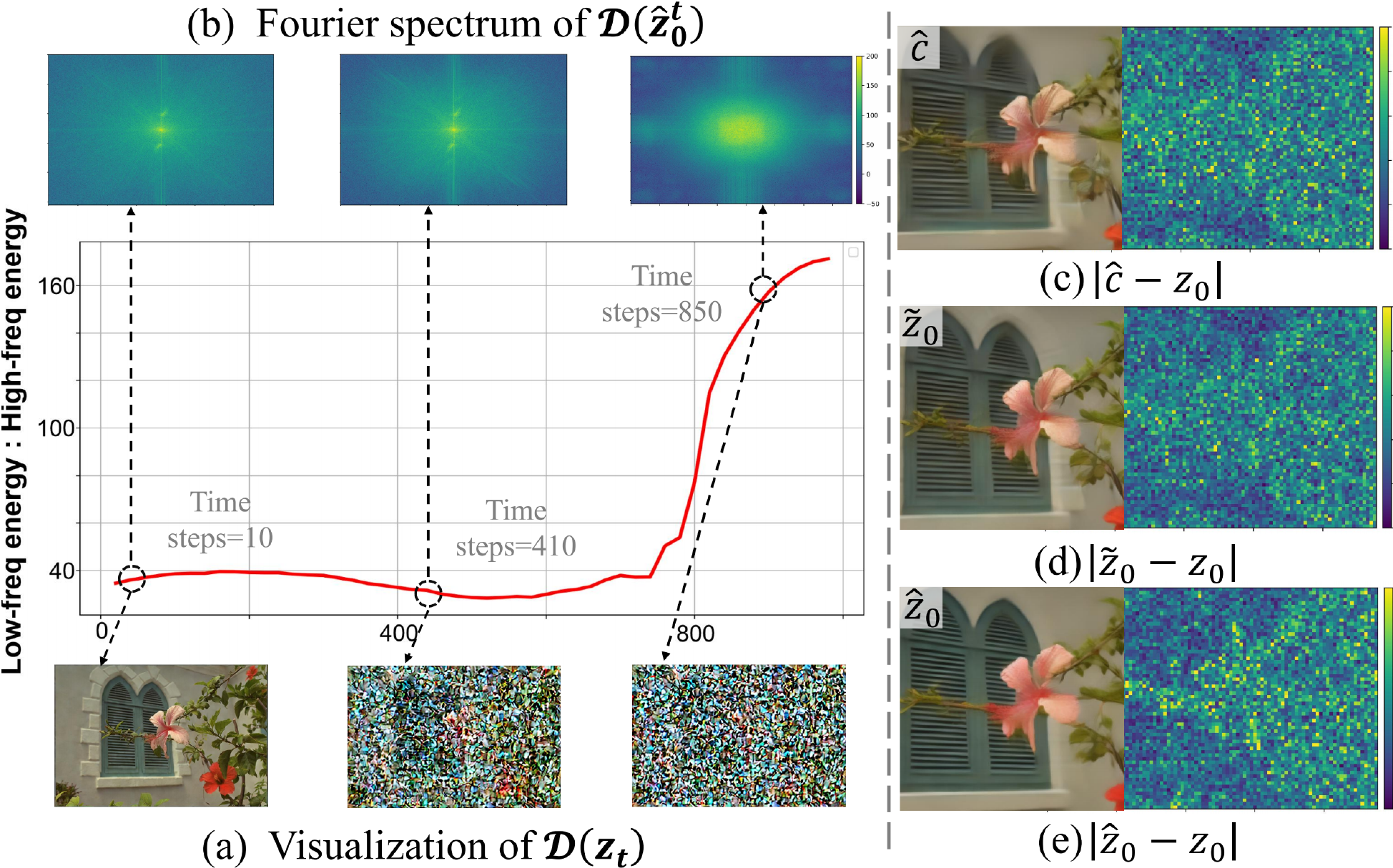}
\caption{(Left) The trend of frequency energy variation in reconstruction by diffusion models at different time steps. 
(Right) Quantitative performance comparison of the compressor, denoising network, and FaSE in proceeding $\boldsymbol{z_0}$-prediction.}
\label{fig:smllpic1}
\end{figure}
\subsection{Consistency Refinement for Latent Compression}
\label{subsec1}

\subsubsection{Frequency-aware Skip Estimation}
The foundational diffusion model optimizes \cref{eq:dloss} by infusing conditions into the denoising model, gradually enhancing the realism of generated images. However, this objective cannot serve as the reconstruction target for the compressor, manifesting as pattern collapse \citep{li2024towards}. 
Instead, existing methods align latent \cref{eq:latent_c} to train the compressor, yet the information in the latent stems from the priors of the encoder-decoder $\mathcal{E}$ and $\mathcal{D}$, not the denoising network $\boldsymbol{\epsilon}_{\theta}$. FaSE aims to establish a consistency estimator $\boldsymbol{f}_{\phi,\theta}(\boldsymbol{z_t},\boldsymbol{\hat{c}},t)$ that transforms the diffusion prior and compressed latent conditions into the same predictive target $\boldsymbol{z_0}$:

\begin{align}
\resizebox{0.9\linewidth}{!}{$
\begin{aligned}
    \boldsymbol{f}_{\phi,\theta}(\boldsymbol{z_t},\boldsymbol{\hat{c}},t)=c_{\mathrm{skip}}(t) \boldsymbol{z_t}+c_{\mathrm{out}}(t)\left(\boldsymbol{F}_{\phi}(\boldsymbol{\epsilon}_{\theta}(\boldsymbol{z_t}),\boldsymbol{\hat{c}},t)\right).
    \end{aligned}$}
\end{align}

\begin{figure}[t]
\centering
\includegraphics[width=1\linewidth]{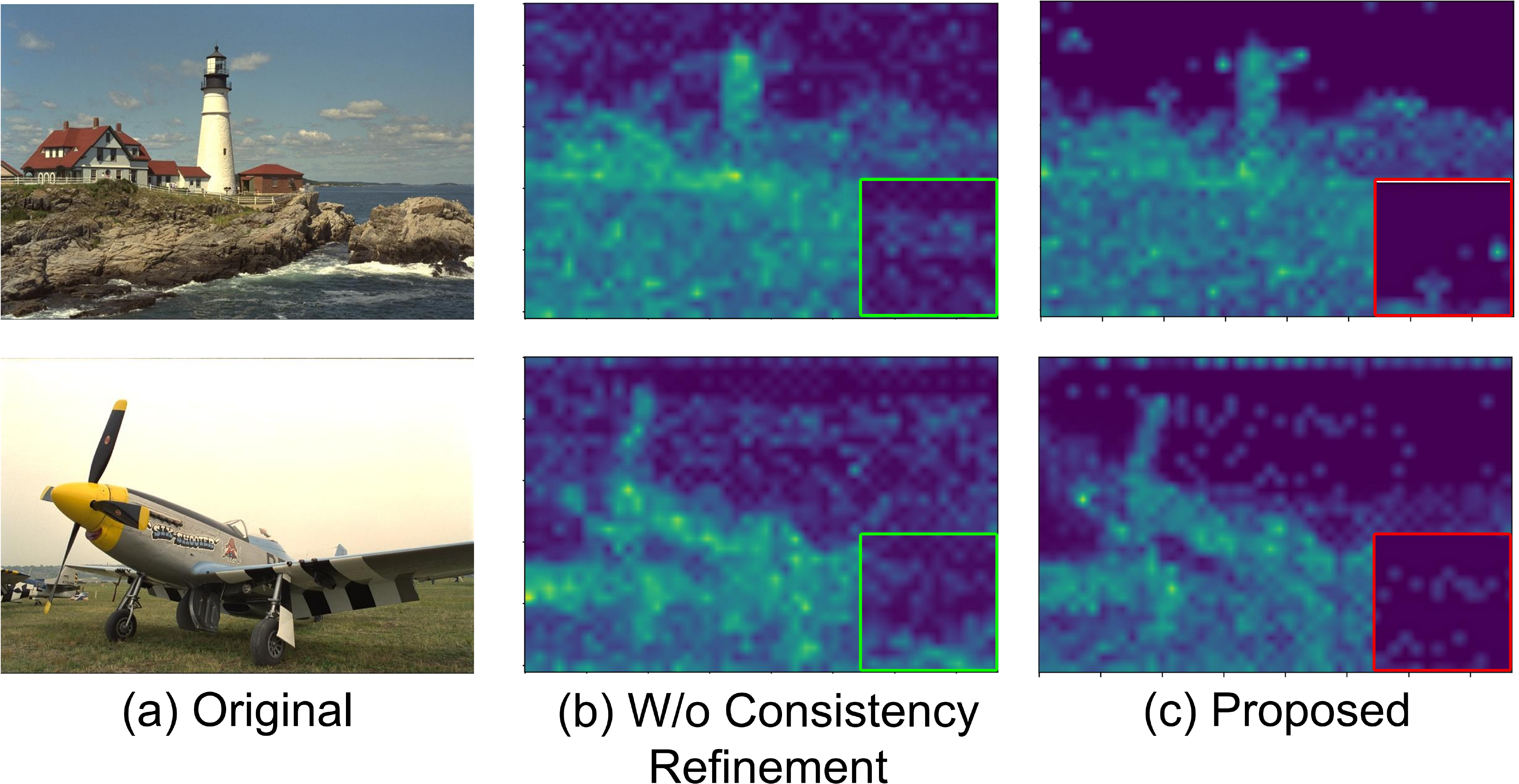}
\caption{Visualization of bit-rate allocation on the Kodak.}
\label{fig:bitallocate}
\end{figure}
\begin{figure*}[t]
\centering
\includegraphics[width=0.9\linewidth]{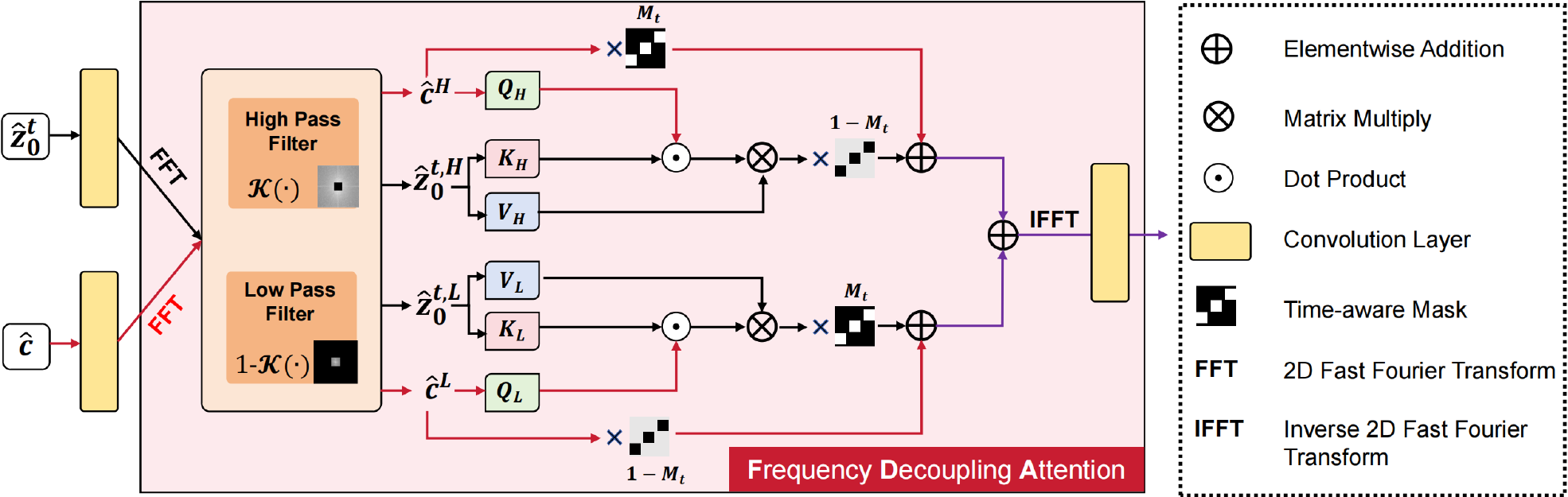}
\caption{Illustration of the FDA module, where $Q_{\cdot}K_{\cdot}V_{\cdot}$ corresponds to Query, Key, and Value in the attention mechanism.}
\label{fig:fda}
\end{figure*}
Where the objective of $\boldsymbol{F}_{\phi}(\cdot,\cdot,\cdot)$ is to align the $\boldsymbol{\epsilon}$-Prediction prior with the image condition $\boldsymbol{\hat{c}}$, predicting the solution of the PF-ODE through consistency loss \citep{lcm}:
\begin{align}
\resizebox{0.9\linewidth}{!}{$
\begin{aligned}
&\mathcal{L}_{\mathcal{C}}\left(\boldsymbol{\phi}, \boldsymbol{\phi}^{-} ; \Psi\right)=
\\
&\mathbb{E}_{\boldsymbol{z}, \boldsymbol{\hat{c}}, n}\left[d\left(\boldsymbol{F}_{\boldsymbol{\phi}}\left(\boldsymbol{\epsilon}_{\theta}(\boldsymbol{z}_{t_{n+k}}), \boldsymbol{\hat{c}}, t_{n+1}\right), \boldsymbol{F}_{\boldsymbol{\phi}^{-}}\left(\boldsymbol{\epsilon}_{\theta}(\hat{\boldsymbol{z}}_{t_n}^{\Psi}),\boldsymbol{\hat{c}}, t_n\right)\right)\right].
\end{aligned}$}
\end{align}
Where $\Psi$ represents the DDIM-solver \citep{DDIM}, $d(\cdot,\cdot)$ denotes the mean squared error, and $k=20$. In contrast to consistency distillation, $\boldsymbol{F}_{\phi}(\cdot,\cdot,\cdot)$ begins with zero initialization, necessitating the $\boldsymbol{z}_0$-Prediction loss to stabilize the prediction process:
\begin{align}
\label{eq:wt}
    \mathcal{L}_{\boldsymbol{F}}=w(t)\cdot\|\boldsymbol{F}_{\phi}(\boldsymbol{\epsilon}_{\theta}(\boldsymbol{z_t}),\boldsymbol{\hat{c}},t)-\boldsymbol{z_0}\|_2^2.
\end{align}
Here, $w(t)=\frac{1}{2}\left(\frac{\bar{\alpha}{t-1}}{1-\bar{\alpha}{t-1}}-\frac{\bar{\alpha}_t}{1-\bar{\alpha}_t}\right)$.

We refer to the predicted $\boldsymbol{z_0}$ obtained through $\boldsymbol{\epsilon}$-Prediction (\cref{eq:eps}) as $\boldsymbol{\hat{z}_0}$, and the prediction derived from FaSE as $\boldsymbol{\tilde{z}_0}$. The right side of \cref{fig:smllpic1} illustrates three distinct patterns of $\boldsymbol{z_0}$ prediction. $\boldsymbol{\hat{c}}$ provides the roughest prediction (\cref{fig:smllpic1}(c)), with the predicted differences evenly distributed throughout the image. Although $\boldsymbol{\hat{z}_0}$ refines many textures using priors (\cref{fig:smllpic1}(e)), it exhibits significant inconsistencies and distortions in specific details. In contrast, FaSE effectively aligns $\boldsymbol{\hat{c}}$ and $\boldsymbol{\hat{z}_0}$, combining detail consistency with texture richness (\cref{fig:smllpic1}(d)). Crucially, the joint training of the compressor with FaSE enables it to actively capture the attention of the diffusion prior in different regions to extract more compact compression representations. As shown in \cref{fig:bitallocate}, compared to not utilizing consistency refinement, the compressor tends to waste bits on flat, highly repetitive patterns in natural images (such as sky, grass, etc.),  as shown in \cref{fig:bitallocate}(b). However, these common textures can be easily synthesized by the diffusion prior. In contrast, FaSE perfectly captures these prior signals, significantly optimizing the bit rate allocation of the compressor.
\subsubsection{Frequency Decoupling Attention }

The signals recovered by the diffusion model at different time steps exhibit non-uniformity in the frequency domain. Specifically, in the early time steps, the denoising network tends to focus on recovering low-frequency signals, while high-frequency signals are synthesized more in later time steps \citep{freq1,freq2}. Our experiments also confirm this observation, as shown on the left side of \cref{fig:smllpic1}. This implies that when designing $\boldsymbol{F}_{\phi}(\cdot,\cdot,\cdot)$, we need to consider the alignment of $\boldsymbol{\hat{z}^{t}_0}$ and $\boldsymbol{\hat{c}}$ in different frequency domains for different time steps $t$. To address this, we introduce Frequency-Decoupling Attention (FDA), as illustrated in \cref{fig:fda}. Specifically, FDA first transforms $\boldsymbol{\hat{z}^{t}_0}$ and $\boldsymbol{\hat{c}}$ to the Fourier domain using fast Fourier transform $FFT(\cdot)$, and then obtains the high and low-frequency components of both through high-pass and low-pass filters:
\begin{align}
\resizebox{0.85\linewidth}{!}{$
\begin{aligned}
&\boldsymbol{\hat{c}^H}=FFT(\boldsymbol{\hat{c}})\odot\mathcal{K}, \boldsymbol{\hat{c}^L}=FFT(\boldsymbol{\hat{c}})\odot(1-\mathcal{K}),
\\
&\boldsymbol{\hat{z}^{t,H}_0}=FFT(\boldsymbol{\hat{z}^{t}_0})\odot\mathcal{K}, \boldsymbol{\hat{z}^{t,L}_0}=FFT(\boldsymbol{\hat{z}^{t}_0})\odot(1-\mathcal{K}).
\end{aligned}$}
\end{align}
We then employ two separate cross-attention mechanisms to individually match the high and low-frequency components:
\begin{align}
\resizebox{0.85\linewidth}{!}{$
\begin{aligned}
&Crossattn^H(\boldsymbol{\hat{z}^{t,H}_0},\boldsymbol{\hat{c}^H})=\operatorname{Softmax}\left(\frac{Q^H K^H}{\sqrt{d}}\right) \cdot V^H,
\\
&Crossattn^L(\boldsymbol{\hat{z}^{t,L}_0},\boldsymbol{\hat{c}^L})=\operatorname{Softmax}\left(\frac{Q^L K^L}{\sqrt{d}}\right) \cdot V^L.
\end{aligned}$}
\end{align}
Subsequently, we modulate the contribution of the prior $\boldsymbol{\hat{z}^{t}_0}$ in the prediction using a linear temporal mask, following best practices and previous research experience, where the proportion of the prior's high-frequency content in the mixture increases gradually as the time step approaches zero. Taking the high-frequency component as an example:
\begin{align}
\resizebox{0.85\linewidth}{!}{$
\begin{aligned}
\boldsymbol{\hat{z}^{t,H,o}_0}=Cattn^H(\boldsymbol{\hat{z}^{t,H}_0},\boldsymbol{\hat{c}^H})*M_t+\boldsymbol{\hat{c}^H}*(1-M_t).
\end{aligned}$}
\end{align}
Where $M_t=1-t/T$, with $T$ being the total number of timesteps in the diffusion model.
We present ablation experiments on this module in \cref{tab:ablation}, demonstrating that compared to directly using cross-attention, FDA achieves significant performance gains through the temporal attention mechanism and frequency decoupling.

\subsection{Control Foundation Diffusion with Hybrid Control}
\label{subsec2}
Inspired by \citep{xiadiffpc}, we integrate image control branch reconstructed by the compressor and semantic control branch extracted by CLIP \citep{clip}. These are injected into the denoising network via control encoder \citep{zhang2023adding} and cross-attention (\cref{fig:pipeline}). The image control enhances edge and texture generation for better fidelity, while the semantic control stabilizes generation and corrects color shifts caused by denoising.

For image control, an end-to-end compressor performs latent-space compression and reconstruction. Feature \(\boldsymbol{z_0}\) is downsampled to \(\boldsymbol{y}\) by \(g_a\). A VQ-based categorical hyperprior model \citep{jia2024generative} extracts and compresses hyperpriors, which are then used with the SCCTX entropy model \citep{he2022elic} for \(\boldsymbol{\hat{y}}\)'s entropy estimation. Finally, \(g_s\) decodes \(\boldsymbol{\hat{y}}\) to obtain image-level control \(\boldsymbol{\hat{c}}\).

For the semantic control, an image captioning model \citep{wang2022ofa} extracts image textual descriptions (transmission cost $\leq 0.0001$ bpp). Two Clip-Encoders extract image and textual semantics separately; these embeddings are modulated via a resampling layer \citep{ipada} and injected into the denoising network. Compared to sole textual control, such multimodal semantics enhance decoding performance without transmission redundancy.
\begin{figure*}[t]
\centering
\includegraphics[width=1\linewidth]{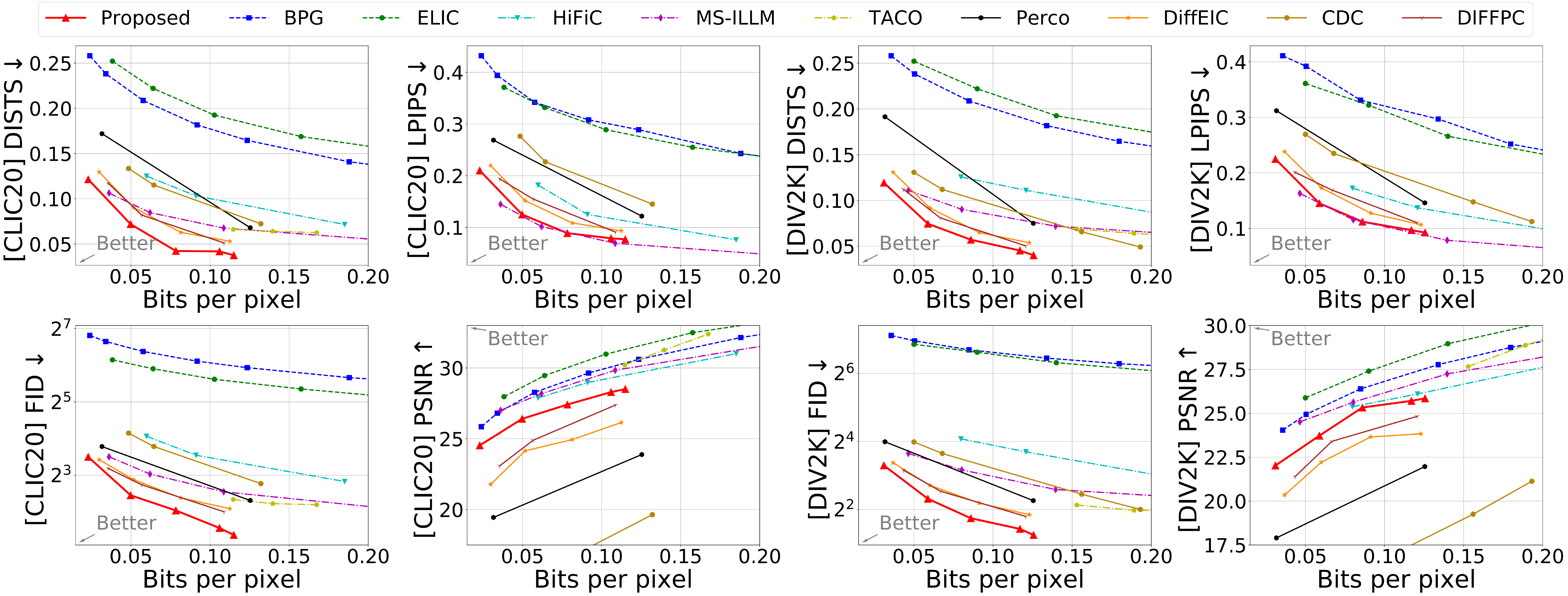}
\caption{Qualitative comparison on the CLIC20 and DIV2K datasets. Grey arrows indicate the direction of better performance.}
\label{fig:rd}
\end{figure*}
\subsection{Training Strategy}
Our model is trained in two stages. In the first stage of training, the compressor, control module, and skip estimation module are jointly trained. While optimizing the compressor's ability to extract representations, the guiding capabilities of the control module and skip estimation are enhanced. For the compressor, we need the rate-distortion loss to balance distortion and compression rate:
\begin{align}
\label{eq:losslc}
    \mathcal{L}_{\text {LC}}=\lambda_1\left\|\boldsymbol{z}_0-\boldsymbol{\hat{c}}\right\|_2^2+\lambda_2R(\hat{\boldsymbol{y}})+\mathcal{L}_{\text {codebook }},
\end{align}
where $\lambda_1$ and $\lambda_2$ are used to balance distortion and bit-rate, and the definition of $\mathcal{L}_{\text {codebook }}$ is:
\begin{align}
\mathcal{L}_{\text {codebook }}=\left\|s g\left(\boldsymbol{l}_y\right)-\hat{\boldsymbol{l}}_y\right\|_2^2+\beta\left\|s g\left(\hat{\boldsymbol{l}}_y\right)-\boldsymbol{l}_y\right\|_2^2,
\end{align}
where $\operatorname{sg}(\cdot)$ denotes the stop-gradient operator and $\beta=$ 0.25.

For the skip estimation module $\boldsymbol{F}_{\phi}(\cdot,\cdot,\cdot)$, we need to maintain its self-consistency while preserving its functionality for $\boldsymbol{z}_0$-Prediction:
\begin{align}
\mathcal{L}_{\text {FaSE}}=\lambda_1\mathcal{L}_{\boldsymbol{F}}+\mathcal{L}_{\mathcal{C}}\left(\boldsymbol{\phi}, \boldsymbol{\phi}^{-} ; \Psi\right).
\end{align}
Since $\mathcal{L}_{\boldsymbol{F}}$ effectively facilitates the optimization of the compressor's distortion, we select hyperparameters that are in line with the distortion settings of the compressor. 

For the control module, we need to optimize the original diffusion loss \cref{eq:dloss} to integrate it into the denoising net:
\begin{align}
\mathcal{L}_{\text{diff}}=\mathbb{E}_{\boldsymbol{z_0},\boldsymbol{\hat{c}},s, t, \epsilon \sim \mathcal{N}(\mathbf{0}, \mathbf{I})}\left\|\boldsymbol{\epsilon}-\boldsymbol{\epsilon}_\theta\left(\boldsymbol{z_t}, \boldsymbol{\hat{c}},s, t\right)\right\|_2^2.
\end{align}
The total loss for the first stage is:
\begin{align}
\label{eq:loss1st}
\mathcal{L}_{\text{1st}}=\mathcal{L}_{\text {LC}}+\mathcal{L}_{\text {FaSE}}+\lambda_3\mathcal{L}_{\text{diff}}.
\end{align}

In the second stage, our main focus is on training $\boldsymbol{F}_{\phi}(\cdot,\cdot,\cdot)$ on the sampling process patterns. We freeze the parameters of the compressor and use $\boldsymbol{F}_{\phi}(\cdot,\cdot,\cdot)$ for \emph{two-step} sampling. We map the sampling results back to the image domain to calculate the reconstruction loss compared to the original image. Additionally, we must ensure that the control module and $\boldsymbol{F}_{\phi}(\cdot,\cdot,\cdot)$ receive proper guidance. Specifically:
\begin{align}
\mathcal{L}_{\text{2st}}=\mathcal{L}_{\text{per}}(\mathcal{D}(\boldsymbol{\hat{z}_0}),\boldsymbol{x})+\mathcal{L}_{\text {FaSE}}+\mathcal{L}_{\text{diff}}.
\end{align}
$\mathcal{L}_{\text{per}}$ represents perceptual loss, and in this case, we are using LPIPS \citep{zhang2018unreasonable}.

\section{Experiments}
\subsection{Experimental Setup}

\subsubsection{Datasets.}
During the validation phase, we referred to \citep{ILLM,CDC} and utilized four widely recognized image compression benchmark datasets: CLIC20 \citep{CLIC2020}, DIV2K \citep{Timofte_2017_CVPR_Workshops}, Kodak \citep{kodak}, and Tecnick \citep{asuni2013testimages}. CLIC20 comprises 428 high-definition images. To ensure more reliable statistical fidelity validation within DIV2K, we selected the initial 400 images from DIV2K-train and 100 images from DIV2K-valid, resulting in a total of 500 images for testing. 

\subsubsection{Baselines.}
We shall compare our proposed DiffCR against a myriad of representative image compression methods, encompassing traditional compression standard such as BPG \citep{bpgbellard}; VAE-based compression method ELIC \citep{he2022elic}; GAN-driven compression techniques including HiFiC \citep{HIFIC}, MS-ILLM \citep{ILLM}, and TACO \citep{lee2024neural}; alongside diffusion-based approaches such as CDC \citep{CDC}, PerCo \citep{perco}, DiffEIC \citep{li2024towards}, DiffPC \citep{xiadiffpc}, and ResULIC \citep{ke2025ultra}. Further details can be perused in the appendix. For certain baselines with undisclosed functionalities \citep{ke2025ultra}, we solely contrasted their publicly available data points on the Kodak.

\subsubsection{Metrics.}
We employ a variety of widely acknowledged metrics to gauge the visual quality of reconstructed images, including perceptual metrics such as LPIPS \citep{zhang2018unreasonable}, DISTS \citep{ding2020image}, FID \citep{heusel2017gans}, and KID \citep{binkowski2018demystifying}. Additionally, we utilize distortion metrics like PSNR and MS-SSIM \citep{ssim} to assess the fidelity of reconstruction. Unless otherwise specified, LPIPS computations are performed using the Alex network, and we also present validation results of LPIPS-V metrics calculated using the VGG network. For the Kodak and Tecnick datasets, due to insufficient data for statistical fidelity metric calculations, we do not validate FID and KID metrics on these datasets. Further details can be found in the appendix.

\subsection{Main Result}
\Cref{fig:rd} illustrates the quantitative comparison between DiffCR and baseline methods on CLIC20 and DIV2K. It is evident that our proposed approach excels in perceptual quality, matching the state-of-the-art solution MS-ILLM in terms of the LPIPS metric and outperforming all diffusion baselines consistently. Particularly, in the appendix, we showcase two metrics, KID and LPIPS-V, demonstrating the consistent superiority of DiffCR over all baselines. Noteworthy is the fact that the enhancement in perceptual metrics does not come at the cost of increased distortion: DiffCR significantly outperforms other diffusion baselines in terms of PSNR, even demonstrating competitiveness with HiFiC on the DIV2K dataset. This clearly indicates that DiffCR achieves a superior distortion-perception trade-off from a compression perspective, rather than solely relying on diffusion model priors for image enhancement. In the supplementary materials, we present qualitative visual comparisons to further elucidate the performance advantages of DiffCR.
\begin{table}
\centering
\begin{tblr}{
  cells = {c},
  cell{1}{1} = {r=2}{},
  cell{1}{2} = {r=2}{},
  cell{1}{3} = {r=2}{},
  vline{4} = {1,2-11}{0.05em},
  hline{1,12} = {-}{0.08em},
  hline{2} = {4}{0.03em},
  hline{3-4,6,11} = {-}{0.05em},
}
Model   & Encoding(s) & Decoding(s) & BD-rate(\%) \\
        &             &             & DISTS       \\
ELIC    & 0.009       & 0.008       & 1017.3     \\
HiFiC   & 0.012       & 0.062       & 376.04      \\
MS-ILLM & 0.069       & 0.068       & 125.25     \\
CDC     & 0.007       & 3.081       & 116.47     \\
Perco   & 0.131       & 7.455       & 90.757      \\
DiffEIC & 0.402       & 6.618       & 28.062      \\
DiffPC  & 0.089       & 7.325       & 26.831      \\
ResULIC & 181.16*     & 1.021~      & 16.904      \\
Ours    & 0.076       & 0.481       & 0           
\end{tblr}
\caption{Encoding and decoding time on Kodak dataset. A higher BD-rate indicates poorer performance. * The values reported in the \citep{ke2025ultra} paper are utilized.}
\label{tab:complex}
\end{table}
\subsubsection{Complexity Comparisons}
\Cref{tab:complex} presents the encoding and decoding latency as well as performance comparison between DiffCR and baseline methods on the Kodak dataset. The performance metric utilized is BD-rate \citep{bd}. Taking DiffCR as the reference model, a higher BD-rate indicates that the model requires more bits to achieve the same metric as the reference. DiffCR not only boasts satisfactory encoding speed but also, due to its requirement for only two sampling steps, exhibits decoding speeds over $10\times$ faster than the typical diffusion baselines that employ 50 steps of sampling and still outperforms the 4-step decoding of \citep{ke2025ultra} by $2\times$.

\subsection{Ablations}
We conducted ablation experiments on the CLIC20, using DiffCR as a baseline to showcase the BD-rate from different ablations, as depicted in \cref{tab:ablation}.

\subsubsection{Consistency Refinement Estimator (CRE).} In \cref{fig:bitallocate}, we demonstrate the advantage of CRE in optimizing bit rate allocation. As shown in \cref{tab:ablation}, removing CRE leads to a performance degradation of up to 40\%.

\subsubsection{Frequency Decoupling Attention (FDA).} We replaced the FDA module with cross-attention layers. As discussed in previous sections, standard cross-attention struggles to capture the intricate context of diffusion priors and compressed latents across different time steps, resulting in a performance decline of approximately 14\%.

\subsubsection{Second Stage Training (2-stage).} Directly training the model in the second stage to learn sampling patterns from image-domain distortion. Omitting this training stage results in a performance drop of around 47\%.

\subsubsection{Semantic Embedding (S.Emb.).} 
Replacing mixed semantic embeddings with pure textual semantic embeddings leads to a 9\% performance decline and introduces slight color deviations in decoded images.

We also discussed the model's sensitivity to text extracted by different image-captioning models and conducted ablation experiments on the sampling steps. Please refer to the appendix for further details.

\begin{table}
\centering
\begin{tblr}{
  cells = {c},
  cell{1}{1} = {r=2}{},
  cell{1}{2} = {r=2}{},
  cell{1}{3} = {r=2}{},
  cell{1}{4} = {r=2}{},
  cell{1}{5} = {c=2}{},
  vline{5} = {1,2-7}{},
  hline{1,8} = {-}{0.08em},
  hline{2} = {5-7}{},
  hline{2,7} = {8}{r},
  hline{3,7} = {-}{0.05em},
}
S.Emb.                        & CRE                           & FDA                           & 2-stage                       & BD-rate(\%) ~ &       \\
                              &                               &                               &                               & LPIPS    & FID   \\
\CheckmarkBold & \CheckmarkBold & \CheckmarkBold &                               & 44.5     & 51.1  \\
\CheckmarkBold &                               &                               & \CheckmarkBold & 38.24    & 41.21 \\
\CheckmarkBold & \CheckmarkBold &                               & \CheckmarkBold & 13.52    & 15.64 \\
                              & \CheckmarkBold & \CheckmarkBold & \CheckmarkBold & 8.32     & 10.11 \\
\CheckmarkBold & \CheckmarkBold & \CheckmarkBold & \CheckmarkBold & 0        & 0     
\end{tblr}
\caption{Ablation studies comparing each module on the
CLIC20 dataset}
\label{tab:ablation}
\end{table}

\section{Conclusions}
In this paper, we introduce DiffCR, a high-fidelity, low-bitrate image compression framework that leverages consistency refinement diffusion priors. Through extensive experimentation, we demonstrate the significant contributions of the frequency-aware skip prediction module in optimizing bit-rate allocation and balancing perception-distortion trade-offs. DiffCR exhibits satisfactory perceptual quality and decoding efficiency in low-bitrate scenarios. Moreover, the flexible model architecture showcases promising prospects for practical applications.

\section{Acknowledgements}
This work is supported in part by the National Natural Science Foundation of China under grant 624B2088, 62301189, 62571298, and Shenzhen Science and Technology Program under Grant KJZD20240903103702004, SYSPG20241211173609009. 
\bibliography{aaai2026}

@article{bpgbellard,
   title = {BPG image format.},
   author = {Bellard, F.},
   journal = {\url{https://bellard.
org/bpg/}},
year={2014}
}

@article{balle17,
  title={End-to-end optimized image compression},
  author={Ball{\'e}, Johannes and Laparra, Valero and Simoncelli, Eero P},
  journal={arXiv preprint arXiv:1611.01704},
  year={2016}
}

@article{balle18,
  title={Variational image compression with a scale hyperprior},
  author={Ball{\'e}, Johannes and Minnen, David and Singh, Saurabh and Hwang, Sung Jin and Johnston, Nick},
  journal={arXiv preprint arXiv:1802.01436},
  year={2018}
}

@article{minnen2018,
  title={Joint autoregressive and hierarchical priors for learned image compression},
  author={Minnen, David and Ball{\'e}, Johannes and Toderici, George D},
  journal={Advances in neural information processing systems},
  volume={31},
  year={2018}
}

@inproceedings{cheng2020,
  title={Learned image compression with discretized gaussian mixture likelihoods and attention modules},
  author={Cheng, Zhengxue and Sun, Heming and Takeuchi, Masaru and Katto, Jiro},
  booktitle={Proceedings of the IEEE/CVF conference on computer vision and pattern recognition},
  pages={7939--7948},
  year={2020}
}

@inproceedings{he2022elic,
  title={Elic: Efficient learned image compression with unevenly grouped space-channel contextual adaptive coding},
  author={He, Dailan and Yang, Ziming and Peng, Weikun and Ma, Rui and Qin, Hongwei and Wang, Yan},
  booktitle={Proceedings of the IEEE/CVF Conference on Computer Vision and Pattern Recognition},
  pages={5718--5727},
  year={2022}
}

@article{bd,
  title={Calculation of average PSNR differences between RD-curves},
  author={Bjontegaard, Gisle},
  journal={ITU SG16 Doc. VCEG-M33},
  year={2001}
}

@inproceedings{freq1,
  title={Boosting diffusion models with moving average sampling in frequency domain},
  author={Qian, Yurui and Cai, Qi and Pan, Yingwei and Li, Yehao and Yao, Ting and Sun, Qibin and Mei, Tao},
  booktitle={Proceedings of the IEEE/CVF conference on computer vision and pattern recognition},
  pages={8911--8920},
  year={2024}
}

@inproceedings{clip,
  title={Learning transferable visual models from natural language supervision},
  author={Radford, Alec and Kim, Jong Wook and Hallacy, Chris and Ramesh, Aditya and Goh, Gabriel and Agarwal, Sandhini and Sastry, Girish and Askell, Amanda and Mishkin, Pamela and Clark, Jack and others},
  booktitle={International conference on machine learning},
  pages={8748--8763},
  year={2021},
  organization={PmLR}
}

@article{freq2,
  title={Freqprior: Improving video diffusion models with frequency filtering gaussian noise},
  author={Yuan, Yunlong and Guo, Yuanfan and Wang, Chunwei and Zhang, Wei and Xu, Hang and Zhang, Li},
  journal={arXiv preprint arXiv:2502.03496},
  year={2025}
}

@inproceedings{blau2019rethinking,
  title={Rethinking lossy compression: The rate-distortion-perception tradeoff},
  author={Blau, Yochai and Michaeli, Tomer},
  booktitle={International Conference on Machine Learning},
  pages={675--685},
  year={2019},
  organization={PMLR}
}

@inproceedings{agustsson2023multi,
  title={Multi-realism image compression with a conditional generator},
  author={Agustsson, Eirikur and Minnen, David and Toderici, George and Mentzer, Fabian},
  booktitle={Proceedings of the IEEE/CVF Conference on Computer Vision and Pattern Recognition},
  pages={22324--22333},
  year={2023}
}

@article{HIFIC,
  title={High-fidelity generative image compression},
  author={Mentzer, Fabian and Toderici, George D and Tschannen, Michael and Agustsson, Eirikur},
  journal={Advances in Neural Information Processing Systems},
  volume={33},
  pages={11913--11924},
  year={2020}
}

@inproceedings{ILLM,
  title={Improving statistical fidelity for neural image compression with implicit local likelihood models},
  author={Muckley, Matthew J and El-Nouby, Alaaeldin and Ullrich, Karen and J{\'e}gou, Herv{\'e} and Verbeek, Jakob},
  booktitle={International Conference on Machine Learning},
  pages={25426--25443},
  year={2023},
  organization={PMLR}
}

@article{CDC,
  title={Lossy image compression with conditional diffusion models},
  author={Yang, Ruihan and Mandt, Stephan},
  journal={Advances in Neural Information Processing Systems},
  volume={36},
  year={2024}
}

@article{hoogeboom2023high,
  title={High-fidelity image compression with score-based generative models},
  author={Hoogeboom, Emiel and Agustsson, Eirikur and Mentzer, Fabian and Versari, Luca and Toderici, George and Theis, Lucas},
  journal={arXiv preprint arXiv:2305.18231},
  year={2023}
}

@article{lei2023text+,
  title={Text+ sketch: Image compression at ultra low rates},
  author={Lei, Eric and Uslu, Yi{\u{g}}it Berkay and Hassani, Hamed and Bidokhti, Shirin Saeedi},
  journal={arXiv preprint arXiv:2307.01944},
  year={2023}
}

@article{wang2024exploiting,
  author = {Wang, Jianyi and Yue, Zongsheng and Zhou, Shangchen and Chan, Kelvin C.K. and Loy, Chen Change},
  title = {Exploiting Diffusion Prior for Real-World Image Super-Resolution},
  journal = {International Journal of Computer Vision},
  year = {2024}
}

@inproceedings{songscore,
  title={Score-Based Generative Modeling through Stochastic Differential Equations},
  author={Song, Yang and Sohl-Dickstein, Jascha and Kingma, Diederik P and Kumar, Abhishek and Ermon, Stefano and Poole, Ben},
year={2021},
  booktitle={International Conference on Learning Representations}
}

@inproceedings{
perco,
title={Towards image compression with perfect realism at ultra-low bitrates},
author={Marlene Careil and Matthew J. Muckley and Jakob Verbeek and St{\'e}phane Lathuili{\`e}re},
booktitle={The Twelfth International Conference on Learning Representations},
year={2024},
url={https://openreview.net/forum?id=ktdETU9JBg}
}

@article{dhariwal2021diffusion,
  title={Diffusion models beat gans on image synthesis},
  author={Dhariwal, Prafulla and Nichol, Alexander},
  journal={Advances in neural information processing systems},
  volume={34},
  pages={8780--8794},
  year={2021}
}

@inproceedings{LDM,
  title={High-resolution image synthesis with latent diffusion models},
  author={Rombach, Robin and Blattmann, Andreas and Lorenz, Dominik and Esser, Patrick and Ommer, Bj{\"o}rn},
  booktitle={Proceedings of the IEEE/CVF conference on computer vision and pattern recognition},
  pages={10684--10695},
  year={2022}
}

@article{DDPM,
  title={Denoising diffusion probabilistic models},
  author={Ho, Jonathan and Jain, Ajay and Abbeel, Pieter},
  journal={Advances in neural information processing systems},
  volume={33},
  pages={6840--6851},
  year={2020}
}

@article{lin2023diffbir,
  title={Diffbir: Towards blind image restoration with generative diffusion prior},
  author={Lin, Xinqi and He, Jingwen and Chen, Ziyan and Lyu, Zhaoyang and Dai, Bo and Yu, Fanghua and Ouyang, Wanli and Qiao, Yu and Dong, Chao},
  journal={arXiv preprint arXiv:2308.15070},
  year={2023}
}

@article{DDIM,
  title={Denoising diffusion implicit models},
  author={Song, Jiaming and Meng, Chenlin and Ermon, Stefano},
  journal={arXiv preprint arXiv:2010.02502},
  year={2020}
}

@inproceedings{wang2018esrgan,
  title={Esrgan: Enhanced super-resolution generative adversarial networks},
  author={Wang, Xintao and Yu, Ke and Wu, Shixiang and Gu, Jinjin and Liu, Yihao and Dong, Chao and Qiao, Yu and Change Loy, Chen},
  booktitle={Proceedings of the European conference on computer vision (ECCV) workshops},
  pages={0--0},
  year={2018}
}

@article{asuni2013testimages,
  title={TESTIMAGES: A large data archive for display and algorithm testing},
  author={Asuni, Nicola and Giachetti, Andrea},
  journal={Journal of Graphics Tools},
  volume={17},
  number={4},
  pages={113--125},
  year={2013},
  publisher={Taylor \& Francis}
}

@article{jpeg,
  title={The JPEG still picture compression standard},
  author={Wallace, Gregory K},
  journal={IEEE transactions on consumer electronics},
  volume={38},
  number={1},
  pages={xviii--xxxiv},
  year={1992},
  publisher={IEEE}
}

@inproceedings{song2023consistency,
  title={Consistency Models},
  author={Song, Yang and Dhariwal, Prafulla and Chen, Mark and Sutskever, Ilya},
  booktitle={International Conference on Machine Learning},
  pages={32211--32252},
  year={2023},
  organization={PMLR}
}

@article{lcm,
  title={Latent consistency models: Synthesizing high-resolution images with few-step inference},
  author={Luo, Simian and Tan, Yiqin and Huang, Longbo and Li, Jian and Zhao, Hang},
  journal={arXiv preprint arXiv:2310.04378},
  year={2023}
}

@article{shannon1948mathematical,
  title={A mathematical theory of communication},
  author={Shannon, Claude Elwood},
  journal={The Bell system technical journal},
  volume={27},
  number={3},
  pages={379--423},
  year={1948},
  publisher={Nokia Bell Labs}
}

@inproceedings{ssim,
  title={Multiscale structural similarity for image quality assessment},
  author={Wang, Zhou and Simoncelli, Eero P and Bovik, Alan C},
  booktitle={The Thrity-Seventh Asilomar Conference on Signals, Systems \& Computers, 2003},
  volume={2},
  pages={1398--1402},
  year={2003},
  organization={Ieee}
}

@inproceedings{zhang2018unreasonable,
  title={The unreasonable effectiveness of deep features as a perceptual metric},
  author={Zhang, Richard and Isola, Phillip and Efros, Alexei A and Shechtman, Eli and Wang, Oliver},
  booktitle={Proceedings of the IEEE conference on computer vision and pattern recognition},
  pages={586--595},
  year={2018}
}

@article{ding2020image,
  title={Image quality assessment: Unifying structure and texture similarity},
  author={Ding, Keyan and Ma, Kede and Wang, Shiqi and Simoncelli, Eero P},
  journal={IEEE transactions on pattern analysis and machine intelligence},
  volume={44},
  number={5},
  pages={2567--2581},
  year={2020},
  publisher={IEEE}
}

@article{heusel2017gans,
  title={Gans trained by a two time-scale update rule converge to a local nash equilibrium},
  author={Heusel, Martin and Ramsauer, Hubert and Unterthiner, Thomas and Nessler, Bernhard and Hochreiter, Sepp},
  journal={Advances in neural information processing systems},
  volume={30},
  year={2017}
}

@article{binkowski2018demystifying,
  title={Demystifying mmd gans},
  author={Bi{\'n}kowski, Miko{\l}aj and Sutherland, Danica J and Arbel, Michael and Gretton, Arthur},
  journal={arXiv preprint arXiv:1801.01401},
  year={2018}
}

@misc{CLIC2020,
  title = {Workshop and Challenge on Learned Image Compression (CLIC2020)},
  author = {George Toderici, Wenzhe Shi},
  url = {http://www.compression.cc},
  year={2020},
  organization={CVPR}
}

@InProceedings{Timofte_2017_CVPR_Workshops,
author = {Timofte, Radu and Agustsson, Eirikur and Van Gool, Luc and Yang, Ming-Hsuan and Zhang, Lei and Lim, Bee and others},
title = {NTIRE 2017 Challenge on Single Image Super-Resolution: Methods and Results},
booktitle = {The IEEE Conference on Computer Vision and Pattern Recognition (CVPR) Workshops},
month = {July},
year = {2017}
}

@misc{kodak,
  title = { Kodak lossless true color image suite},
  author = {Eastman Kodak Company},
  year={2013},
  url = {http://r0k.us/graphics/kodak/},
}

@inproceedings{jia2024generative,
  title={Generative latent coding for ultra-low bitrate image compression},
  author={Jia, Zhaoyang and Li, Jiahao and Li, Bin and Li, Houqiang and Lu, Yan},
  booktitle={Proceedings of the IEEE/CVF Conference on Computer Vision and Pattern Recognition},
  pages={26088--26098},
  year={2024}
}

@inproceedings{mao2024extreme,
  title={Extreme image compression using fine-tuned vqgans},
  author={Mao, Qi and Yang, Tinghan and Zhang, Yinuo and Wang, Zijian and Wang, Meng and Wang, Shiqi and Jin, Libiao and Ma, Siwei},
  booktitle={2024 Data Compression Conference (DCC)},
  pages={203--212},
  year={2024},
  organization={IEEE}
}

@article{li2024towards,
  author={Li, Zhiyuan and Zhou, Yanhui and Wei, Hao and Ge, Chenyang and Jiang, Jingwen},
  journal={IEEE Transactions on Circuits and Systems for Video Technology}, 
  title={Towards Extreme Image Compression with Latent Feature Guidance and Diffusion Prior}, 
  year={2024},
  doi={10.1109/TCSVT.2024.3455576}}

@inproceedings{lu2024hybridflow,
  title={HybridFlow: Infusing Continuity into Masked Codebook for Extreme Low-Bitrate Image Compression},
  author={Lu, Lei and Xie, Yanyue and Jiang, Wei and Wang, Wei and Lin, Xue and Wang, Yanzhi},
  booktitle={Proceedings of the 32nd ACM International Conference on Multimedia},
  pages={3010--3018},
  year={2024}
}

@inproceedings{lee2024neural,
        title={Neural Image Compression with Text-guided Encoding for both Pixel-level and Perceptual Fidelity},
        author={Lee, Hagyeong and Kim, Minkyu and Kim, Jun-Hyuk and Kim, Seungeon and Oh, Dokwan and Lee, Jaeho},
        booktitle={International Conference on Machine Learning},
        year={2024}
      }

@article{ipada,
  title={Ip-adapter: Text compatible image prompt adapter for text-to-image diffusion models},
  author={Ye, Hu and Zhang, Jun and Liu, Sibo and Han, Xiao and Yang, Wei},
  journal={arXiv preprint arXiv:2308.06721},
  year={2023}
}

@article{wang2022ofa,
  author    = {Peng Wang and
               An Yang and
               Rui Men and
               Junyang Lin and
               Shuai Bai and
               Zhikang Li and
               Jianxin Ma and
               Chang Zhou and
               Jingren Zhou and
               Hongxia Yang},
  title     = {OFA: Unifying Architectures, Tasks, and Modalities Through a Simple Sequence-to-Sequence
               Learning Framework},
  journal   = {CoRR},
  volume    = {abs/2202.03052},
  year      = {2022}
}

@inproceedings{xiadiffpc,
  title={DiffPC: Diffusion-based High Perceptual Fidelity Image Compression with Semantic Refinement},
  author={Xia, Yichong and Zhou, Yimin and Wang, Jinpeng and An, Baoyi and Wang, Haoqian and Wang, Yaowei and Chen, Bin},
  year={2025},
  booktitle={The Thirteenth International Conference on Learning Representations}
}

@article{ke2025ultra,
  title={Ultra Lowrate Image Compression with Semantic Residual Coding and Compression-aware Diffusion},
  author={Ke, Anle and Zhang, Xu and Chen, Tong and Lu, Ming and Zhou, Chao and Gu, Jiawen and Ma, Zhan},
  journal={arXiv preprint arXiv:2505.08281},
  year={2025}
}

@inproceedings{zhang2023adding,
  title={Adding conditional control to text-to-image diffusion models},
  author={Zhang, Lvmin and Rao, Anyi and Agrawala, Maneesh},
  booktitle={Proceedings of the IEEE/CVF international conference on computer vision},
  pages={3836--3847},
  year={2023}
}

@inproceedings{liuexploration,
  title={An Exploration with Entropy Constrained 3D Gaussians for 2D Video Compression},
year={2025},
  author={Liu, Xiang and Chen, Bin and Liu, Zimo and Wang, Yaowei and Xia, Shu-Tao},
  booktitle={The Thirteenth International Conference on Learning Representations}
}

@article{liu2025efficient,
  title={An efficient implicit neural representation image codec based on mixed autoregressive model for low-complexity decoding},
  author={Liu, Xiang and Chen, Jiahong and Chen, Bin and Liu, Zimo and An, Baoyi and Xia, Shu-Tao and Wang, Zhi},
  journal={IEEE Transactions on Multimedia},
  year={2025},
  publisher={IEEE}
}

@article{qin2024mambavc,
  title={Mambavc: Learned visual compression with selective state spaces},
  author={Qin, Shiyu and Wang, Jinpeng and Zhou, Yimin and Chen, Bin and Luo, Tianci and An, Baoyi and Dai, Tao and Xia, Shutao and Wang, Yaowei},
  journal={arXiv preprint arXiv:2405.15413},
  year={2024}
}

@article{qin2023perceptual,
  title={Perceptual image compression with cooperative cross-modal side information},
  author={Qin, Shiyu and Chen, Bin and Huang, Yujun and An, Baoyi and Dai, Tao and Xia, Shu-Tao},
  journal={arXiv preprint arXiv:2311.13847},
  year={2023}
}

@article{li2024diffusion,
  title={Diffusion-based Extreme Image Compression with Compressed Feature Initialization},
  author={Li, Zhiyuan and Zhou, Yanhui and Wei, Hao and Ge, Chenyang and Mian, Ajmal},
  journal={arXiv preprint arXiv:2410.02640},
  year={2024}
}

\end{document}